\documentclass[conference]{IEEEtran}
\makeatletter
\def\ps@headings{%
\def\@oddhead{\mbox{}\scriptsize\rightmark \hfil \thepage}%
\def\@evenhead{\scriptsize\thepage \hfil \leftmark\mbox{}}%
\def\@oddfoot{}%
\def\@evenfoot{}}
\makeatother
\usepackage[justification=centering]{caption}
\usepackage{url}
\usepackage{booktabs}
\usepackage{graphicx,subfigure}
\usepackage{epstopdf}
\usepackage{amsmath}
\usepackage{algorithm}
\usepackage{algpseudocode}
\usepackage{amsmath}
\usepackage{amssymb}
\usepackage{amsthm}
\usepackage{epsfig}

\usepackage{amsfonts}
\usepackage{multirow}
\usepackage{color}
\usepackage{array}
\usepackage{listings}
\usepackage{hyperref}
\usepackage[underline=true]{pgf-umlsd}

\usepackage{setspace}

\begin{document}

\title{Histopathologic Cancer Detection}

\author{
  \IEEEauthorblockN{Varan Singh Rohila, Neeraj Lalwani, and Lochan Basyal}
  \IEEEauthorblockA{
    \textit{Stevens Institute of Technology, Hoboken, U.S.A.} \\
    \{vrohila, nlalwan1, lbasyal\}@stevens.edu
  }
}

\maketitle

\begin{abstract}
Early diagnosis of the cancer cells is necessary for making an effective treatment plan and for the health and safety of a patient. Nowadays, doctors usually use a histological grade that pathologists determine by performing a semi-quantitative analysis of the histopathological and cytological features of hematoxylin-eosin (HE) stained histopathological images. This research contributes a potential classification model for cancer prognosis to efficiently utilize the valuable information underlying the HE-stained histopathological images. This work uses the PatchCamelyon benchmark datasets and trains them in a multi-layer perceptron and convolution model to observe the model's performance in terms of precision, Recall, F1 Score, Accuracy, and AUC Score. The evaluation result shows that the baseline convolution model outperforms the baseline MLP model. Also, this paper introduced ResNet50 and InceptionNet models with data augmentation, where ResNet50 is able to beat the state-of-the-art model. Furthermore, the majority vote and concatenation ensemble were evaluated and provided the future direction of using transfer learning and segmentation to understand the specific features. 
\end{abstract}
\section{Introduction}
Cancer has become one of the major health concerns worldwide and early diagnosis of cancer cells plays a vital role in the effective treatment planning of the patient. Cancer screening using breast tissue biopsies aims to distinguish between benign and malignant lesions. However, manual assessment of large-scale histopathological images is a challenging task due to the variance in appearance, heterogeneous structure, and textures\cite{intro1}. Such a manual analysis is laborious, time-consuming, and often dependent on subjective human interpretation. For this reason, the concept of cancer detection with the analysis of histopathological images with machine learning algorithms provides a significant direction in the research of early cancer diagnosis. 

In recent years, deep learning outperformed state-of-the-art methods in machine learning and medical image analysis tasks, including classification, detection, segmentation, and computer-based diagnosis. The PatchCamelyon benchmark dataset as histopathological images has been used for model training and further computation in this research project. These datasets' images are extracted from the histopathologic scans of lymph nose sections. For implementing the machine learning model, this paper mainly focuses on two approaches: a multi-layer-perceptron model and a simple convolutional model for image classification. The model's performance is observed with the Precision, Recall, F1 Score, Accuracy, and AUC Score. 

The rest of the paper is organized so that related work has been discussed in section II. Furthermore, the proposed solution includes the description of the dataset, machine learning algorithms, and implementation details presented in section III. Similarly, the model comparison and the future direction are discussed in IV and V. The purpose of this research is concluded in section VI.

\begin{figure*}[ht!]
\centerline{\includegraphics[width=0.8\textwidth]{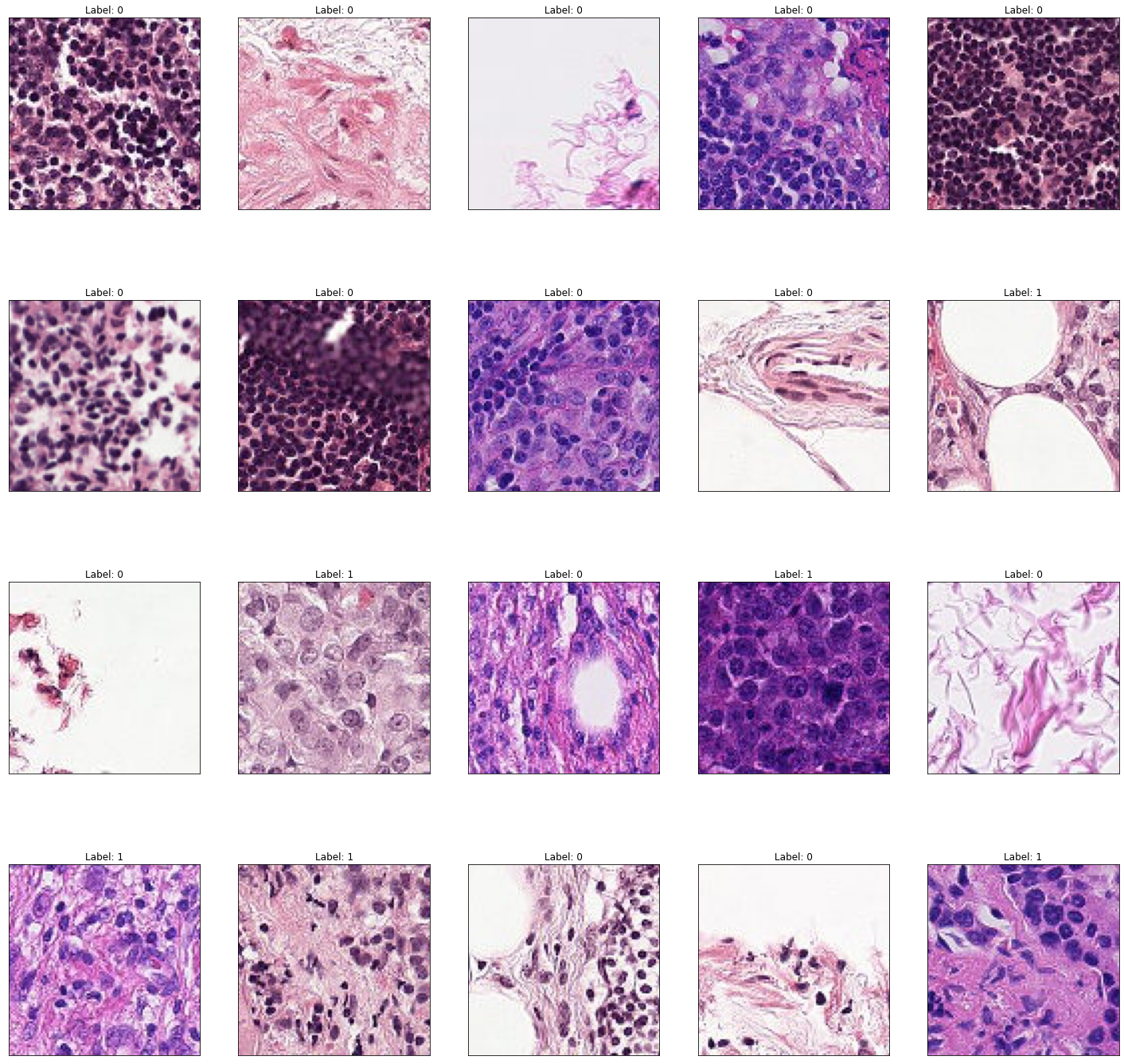}}
\caption{Examples From the Dataset.}
\label{examples}
\end{figure*}

\section{Related Work}
This section summarizes existing works in the field of histopathology cancer detection and image classification in general.

\subsection{Histopathology Cancer Detection}

Various image classification approaches have been proposed for automatic cancer detection. The authors of\cite{tl} presented the concept of transfer learning and deep feature extraction. The two models, AlexNet and Vgg16, are considered for feature extraction, and AlexNet is used for further fine-tuning. Furthermore, a support vector machine(SVM) is used for the classification of the images into benign and malignant classes. This research used the BreaKHis dataset for the experiment and calculated the accuracy scores for performance evaluation.

Other authors\cite{ddtl} proposed an ensemble transfer learning (ETL) framework for classifying well-differentiated, moderately differentiated, and poorly differentiated cervical histopathological images. First, the author developed a transfer learning structure based on Inception V3, Exception, VGG16, and Resnet 50. Then, an ensemble learning strategy based on weighted voting was introduced to improve classification performance. This research claims that the problem of histopathological cancer detection can be solved by adopting the concept of deep multiple instances learning that integrates deep convolutional neural networks and multi-instance learning. 

Continuing with the ensemble techniques, the authors of\cite{ensemble} proposed an ensemble model that adapts three pre-trained CNNs, namely VGG19, MobileNet, and DenseNet. The ensemble model was used for the feature extraction, and a multi-layer perceptron classifier was used to perform the classification task. Furthermore, diverse pre-processing and CNN tuning techniques, stain-normalization, data augmentation, hyperparameter tuning, and fine-tuning were used to train the model. The author of this research validated four publicly available benchmark datasets, i;e., ICIAR, BreakHis, Patch-Cameleon, and Bioimaging.

Lastly, this study\cite{ensemble_two} suggests a deep ensemble model for the binary classification of breast histopathology pictures of benign and malignant lesions based on image-level labeling. The training, validation, and test sets of the BreaKHis dataset are randomly assigned. The numbers of benign and cancerous samples are then balanced using data augmentation techniques. VGG16, Xception, ResNet50, and DenseNet201 are chosen as base classifiers based on their performance in transfer learning and the complementarity between networks. Image-level binary classification in the ensemble network model is accurate to 98.90\%. This approach is experimentally evaluated on the same dataset with the most recent MLP models to confirm its capabilities. In classification tasks, the Ensemble model has a 5-20\% advantage, highlighting its extensive capabilities. Just as a few doctors should be referred to before reaching any conclusion, these works showcase the importance of ensemble learning in combining the prowess of different models.

To understand the scope and landscape of the issue at hand, this paper\cite{reviewpaper} reviewed the detection methods of histopathological cancer cells and predicted the future development trends to guide follow-up research. The paper focuses on machine vision overcomes the disadvantages of traditional detection methods in cancer detection and can help pathologists improve detection accuracy. Furthermore, the author presented the workflow of the machine vision detection system, image acquisition, pre-processing techniques, segmentation, feature extraction, and classification. The system depicted the microscope-mounted digital cameras or scanners used to obtain histopathological images. The pre-processing of the images can be performed through enhancement and color normalization techniques. The author reviewed the various segmentation techniques, threshold segmentation, active contour, clustering, and watershed. After segmentation, feature extraction can be performed with shape features, HSV, and Gray symbiotic. In the last step of the machine vision detection system, the author discussed the supervised and unsupervised classification for finding Benign and Malignant. At every stage, a couple of unique techniques can be applied for better overall results.

\subsection{Image Classification}
Image classification refers to extracting useful information from an image and then classifying it based on certain attributes. The most widely common image classification task is the classification of cats and dogs. Earlier versions of image classification models found even such trivial problems challenging. The introduction of convolution operations and deep neural networks advanced state-of-the-art image classification. Today, new architectures are evaluated on the ImageNet dataset \cite{imagenet}. It has 1000 categories for image classification. Recent state-of-the-art image classification models are:
\begin{itemize}
    \item \textbf{ResNet} Deep convolution models have the issue of vanishing gradients where the gradient becomes so small during backpropagation and weights aren't updated. To counter that issue, authors of \cite{resnet} introduce skip connections in their network, which adds the information from previous layers to the current layer, thus reducing the problem of vanishing gradient descent.
    \item \textbf{DenseNet} Inspired by the skip connections introduced in \cite{resnet}, authors of \cite{densenet} use those skip connections by concatenation the previous layer with the current layer. This has an increased performance from the addition of skip layers of Resnet.
    \item \textbf{InceptionNet} Complexity of convolution models increases as the number of filters increases. The authors of \cite{inceptionnet} introduce a 1x1 convolution operation that helps change the network's filter size. This reduces complexity and offers greater control of the architecture.
    \item \textbf{ViT or Vision Transformer} Transformer models have proven successful in Natural language problems. The authors of \cite{vit} utilize the transformer model in image classification. To create the input embeddings, they divide the image into small image patches, project them in linear space, and then encode their position embeddings.  
\end{itemize}

\section{Our Solution}

\subsection{Description of Dataset}
The dataset we considered to evaluate our approach is the PatchCamelyon benchmark dataset \cite{pacm16}. It consists of 220,025 color images (96 x 96px) extracted from histopathologic scans of lymph node sections. Each image is annotated with a binary label indicating the presence of metastatic tissue. A positive label indicates that the center 32x32px region of a patch contains at least one pixel of tumor tissue. Tumor tissue in the outer region of the patch does not influence the label. The images from the data are shown in \ref{examples}. The dataset is divided into the train, and test sets, with the test set being 25\% of the dataset. The data statistics are tabulated in Table \ref{tab:data}.

\begin{table}[h]
\centering
\begin{tabular}{ccc}
\toprule
\textbf{Statistics} & \textbf{Train Set} & \textbf{Test Set} \\
\hline
Positive Labels & 66,837 &  22,280\\
Negative Labels & 98,181 & 32,727\\
Total & 165,018 & 55,007\\
\end{tabular}
\caption{Dataset Statistics}
\label{tab:data}
\end{table}

\begin{table*}[t!]
\centering
\begin{tabular}{cccccc}
\toprule
\textbf{Models} & \textbf{Precision} & \textbf{Recall} & \textbf{F1 Score} & \textbf{Accuracy} & \textbf{AUC Score}\\
\hline
Baseline MLP & 0.677 & 0.567 & 0.617 & 0.715 & 0.774\\
Baseline Convolution & 0.791 & 0.731 & 0.760 & 0.812 & 0.875\\
ResNet50 & \textbf{0.950} & \textbf{0.932} & \textbf{0.941} & \textbf{0.952} & \textbf{0.988}\\
InceptionNet & 0.929 & 0.920 & 0.925 & 0.925 & 0.983\\
Majority Vote & 0.905 & 0.969 & 0.936 & 0.946 & -\\
Concatenation Ensemble & 0.922 & 0.927 & 0.925 & 0.939 & 0.982\\
\hline
Model proposed in \cite{ensemble}  & 0.957 & 0.952 & 0.955 & 0.946 & -\\
\end{tabular}
\caption{Baseline Results}
\label{tab:baseline_results}
\end{table*}

\subsection{Machine Learning Algorithms}
For our baseline models, we consider a multi-layer perceptron model and a simple convolutional model for image classification. The input of both these models is (96px, 96px, 3), where 3 corresponds to the RGB image channels. Briefly describing the baseline models below:
\begin{itemize}
    \item \textbf{Multi-layer Perceptron Model}: Simple ANN models have been proven to perform well on image tasks such as the MNIST Handwriting Benchmark Dataset. Hence, we consider a single-layer MLP model with 768 hidden nodes with relu activation. A single node with sigmoid activation outputs the predictions.
    \item \textbf{Convolution Model}: A convolution is the simple application of a filter to an input that results in activation. Repeated application of the same filter to an input results in a map of activations called a feature map, indicating the locations and strength of a detected feature in an input, such as an image. We consider a single-layer convolutional model with 32 filters, (3, 3) filter size with relu activation. After that, Max Pooling is applied with (2, 2) filter size. After flattening and adding a single output node with sigmoid activation, the output is calculated.
    \item \textbf{ResNet50 Model}: We utilize data augmentation with the ResNet50 model. To the output of the ResNet50 model, we add a Global Max Pooling layer and a Global Average Pooling Layer. Finally, these three outputs are concatenated and attached to a single output node. A dropout layer of 0.2 is also added to the model.
    \item \textbf{Inception Model}: Similar to the ResNet50 model, inception model is trained and evaluated.
    \item \textbf{Majority Vote Ensemble Model}: We combine the ResNet50 and Inception model with majority voting. This is a hard ensemble.
    \item \textbf{Concatenation Ensemble Model}: We combine the ResNet50 and Inception models using concatenation and train them together as a joint Neural Network. This is a soft ensemble.
\end{itemize}

\begin{figure}
\centerline{\includegraphics[width=\linewidth]{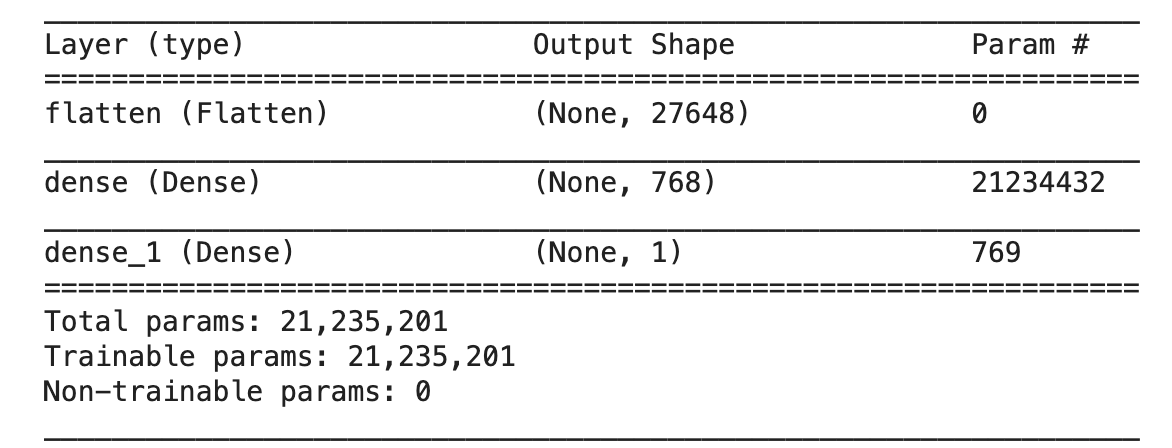}}
\caption{MLP Baseline Model.}
\label{mlp}
\end{figure}

\begin{figure}
\centerline{\includegraphics[width=\linewidth]{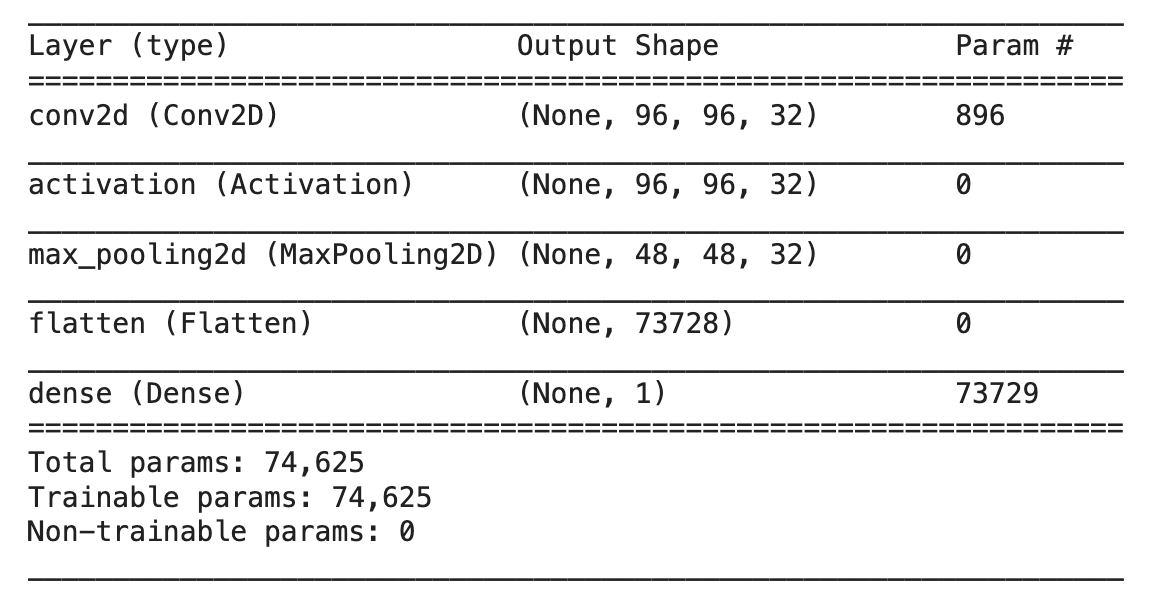}}
\caption{Convolution Baseline Model.}
\label{conv}
\end{figure}

The number of parameters of both the models is shown in figure \ref{mlp} and figure \ref{conv}, respectively.

\begin{figure}
\centerline{\includegraphics[width=\linewidth]{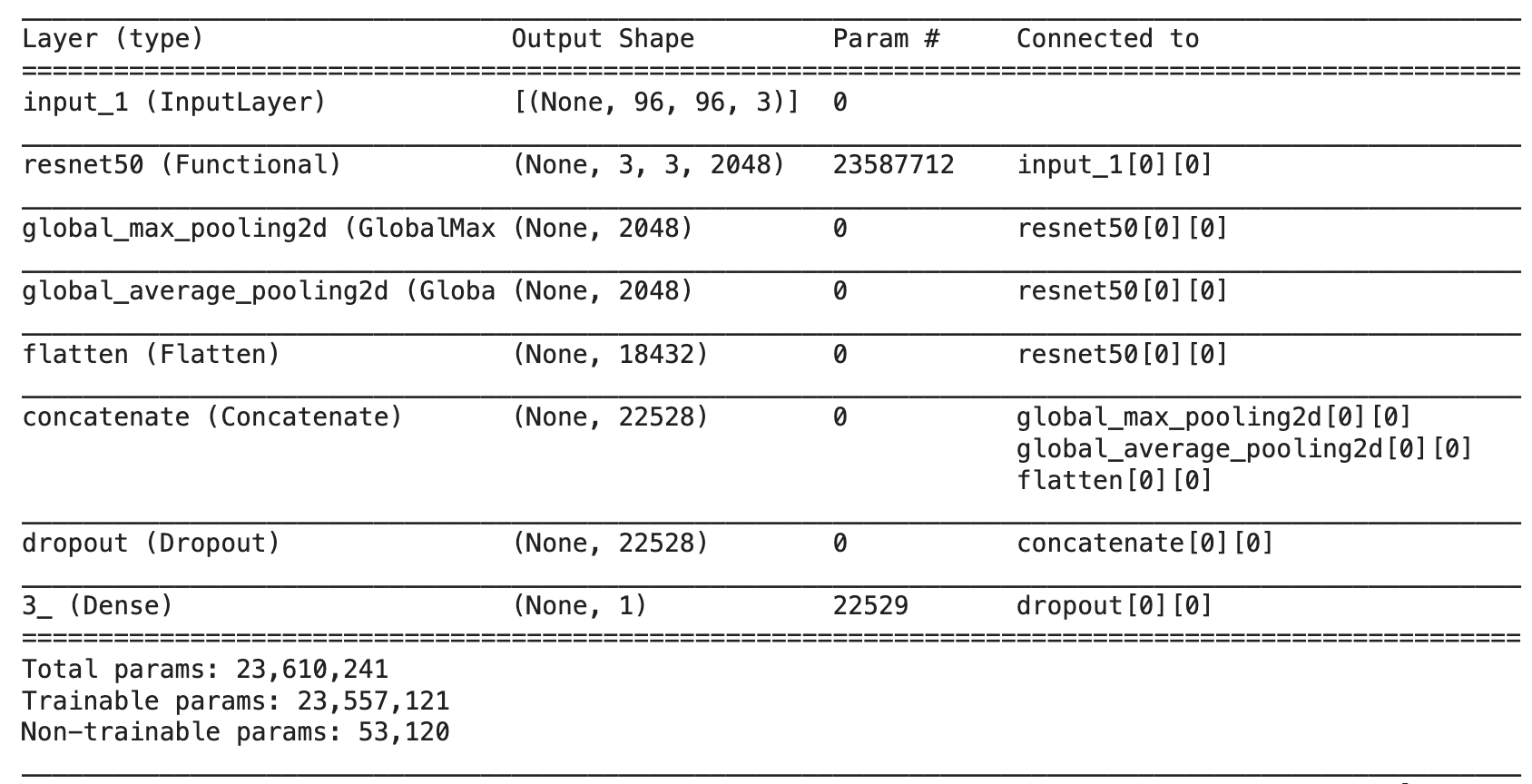}}
\caption{ResNet50 Model.}
\label{resnet}
\end{figure}

\begin{figure}
\centerline{\includegraphics[width=\linewidth]{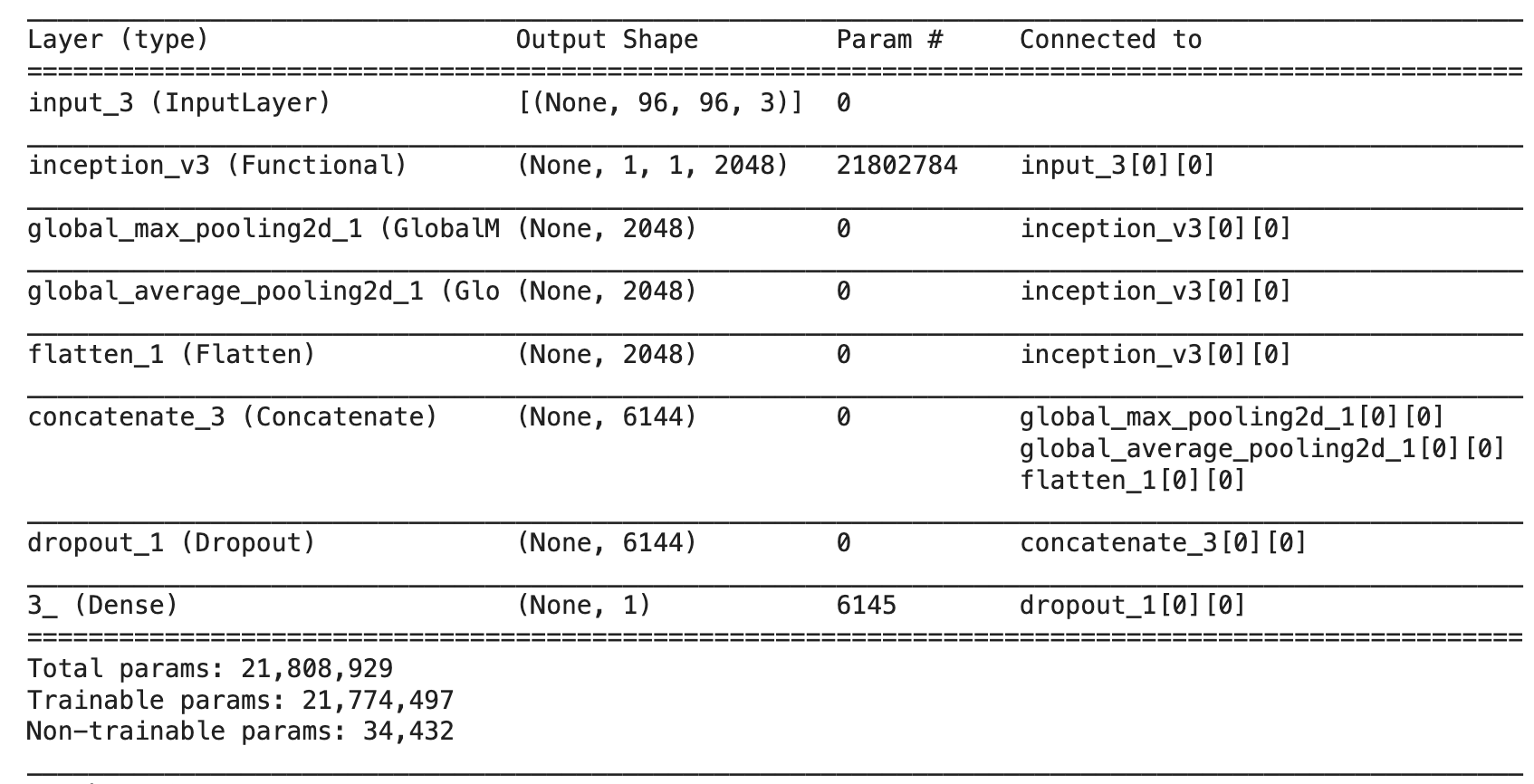}}
\caption{InceptionNet Model.}
\label{inception}
\end{figure}

\subsection{Implementation Details}
The models are trained on the train set for 50 epochs and then evaluated on the test set. Adam is the optimizer for both models, with a 0.0005 learning rate for the baseline MLP and 0.001 for the baseline convolution model. For the ResNet50, the Inception model, and the ensemble model, are trained with a 0.00003 learning rate. Early stopping with patience 5 is used to avoid overfitting.

For data augmentation, random vertical and horizontal flips are applied to the training data with a re-scale of 1/255.

\section{Comparison}
The models are tested on these metrics:

\begin{itemize}
    \item \textbf{Precision}: Precision refers to the number of true positives divided by the total number of positive predictions.
    \item \textbf{Recall}: The recall is the measure of our model correctly identifying True Positives.
    \item \textbf{F1 Score}: It is the harmonic mean of Recall and Precision.
    \item \textbf{Accuracy}: Accuracy is the fraction of predictions our model got right.
    \item \textbf{AUC Score}: AUC score represents the degree or measure of separability.
\end{itemize}

The evaluation results are tabulated in table \ref{tab:baseline_results}. The baseline convolution model outperforms the baseline MLP model in all metrics confirming the prowess of simple convolution networks in image classification. Moreover, due to the weight-sharing property of convolutions, the number of trainable parameters is considerably less than in the baseline MLP model.

\section{Future Directions}
In this work, we evaluate the performances of baseline models on the Patch-Cameleyon dataset. The classification performances can be further increased by using the following:
\begin{itemize}
    \item \textbf{Transfer Learning}: Transfer learning refers to the technique of applying already learned knowledge from one application area to another with shared weights and retraining some layers of the model.
    \item \textbf{Segmentation}: In medical image applications, segmentation extracts useful information about the subject, i.e., the tissue, which can help the model to specific features and ensures the model's focus.
\end{itemize}

\section{Conclusion}
Automatic detection of cancer tissues can help doctors identify tumor tissues and act accordingly. This paper proposes and evaluates baseline models for detecting cancer tissue in lymph nodes. The baseline convolution model achieves an accuracy of 81.2\%, which is quite good for a single-layer model indicating bigger, better models equipped with other techniques such as data augmentation, transfer learning,  and segmentation can achieve state-of-the-art performance on this benchmark dataset. As we advance with that idea, ResNet50 and InceptionNet models were trained with data augmentation. Finally, their majority and Concatenation ensemble were also evaluated on the dataset. Extensive experimentation showcases that the ResNet50 model trained with proper data augmentation is able to beat the state-of-the-art model. In the future, transfer learning and segmentation techniques can be utilized to create a more general and efficient model.

\bibliographystyle{IEEEtran}

\end{document}